\documentclass[journal]{IEEEtran}
\usepackage{amsmath}
\usepackage{amssymb}
\usepackage{hyperref}

\ifCLASSINFOpdf
\else
\fi

\begin{document}
%
\title{Emotion Detection Using Conditional Generative Adversarial Networks (cGAN): A Deep Learning Approach\\}
%
%
%

\author{Anushka~Srivastava ~\IEEEmembership{ Student}}

\maketitle

\begin{abstract}
Emotion recognition is a key task in affective computing with applications in healthcare, human-computer interaction, and surveillance systems. Traditional supervised learning models often struggle with limited and imbalanced datasets, particularly in facial emotion recognition. This study proposes a Conditional Generative Adversarial Network (cGAN)-based approach to generate synthetic emotion-specific facial images to augment training data and mitigate class imbalance. The generator learns to synthesize grayscale $64 \times 64$ facial images conditioned on emotion labels, while the discriminator distinguishes between real and generated images using label conditioning. The model was trained on the FER-2013 dataset and evaluated over 300 epochs. Training results demonstrate stable adversarial loss convergence, indicating effective learning and generation capability. Visual inspection confirms that the generated samples exhibit distinct features corresponding to various emotions. This work highlights the potential of cGANs in improving emotion recognition pipelines by enhancing data diversity and representation for underrepresented emotion classes.
\end{abstract}


%
\IEEEpeerreviewmaketitle

\section{Introduction}
%
%
%
%
\IEEEPARstart{E}{motion} detection is a fundamental component of affective computing, with wide-ranging applications in human-computer interaction, mental health monitoring, and intelligent virtual assistants.\cite{madaan2024multimodal} Emotion detection plays a critical role in enabling machines to understand and respond to human emotions, which is essential for applications in healthcare, human-computer interaction, customer service, and surveillance systems. Traditional approaches to emotion recognition rely on rule-based systems or supervised learning using handcrafted features.\cite{pantic2003affect} However, these methods often struggle with generalization due to limited training data and the complex, subjective nature of emotions.
In the last few years, there have been huge advances in the field of Artificial Intelligence and Deep Learning. These advancements improved the performance of Emotion Detection. Among these, Convolutional Neural Networks (CNNs) and Recurrent Neural Networks (RNNs) have outperformed the most.\\
However, these models remain data hungry and sensitive to imbalanced datasets, often failing to generate robust performance across diverse demographics and real-world conditions.\\
GANs, introduced by Ian Goodfellow in his 2014 paper Generative Adversarial Nets, offer a groundbreaking solution to this challenge. This innovative framework has transformed generative modeling, making it easier to develop models and algorithms capable of creating high-quality, realistic data.\cite{ibmGAN}A Conditional GAN is a type of generative adversarial network that includes additional information, called "labels" or "conditions".\\
This research explores the use of Conditional GANs for augmenting emotion datasets and improving emotion classification performance. We hypothesize that the integration of cGAN-generated synthetic samples can enhance emotion recognition accuracy, particularly in imbalanced or small datasets. The study focuses on evaluating cGAN effectiveness in both data generation quality and downstream classification improvements.


\subsection{Proposed Methodology}
This section explains the complete workflow for Emotion Detection using Conditional Generative Adversarial Network(cGANs).

\subsubsection{Dataset Collection and Prepossessing}
The dataset used for the proposed system is cleaned and preprocessed version of the FER-2013 (Facial Expression Recognition 2013). The data set contains grayscaled images of $64 \times 64$ pixels resolution.The total data set contains 28,709 images belonging to seven different classes.The corresponding label array shaped as $(28709,)$ and image array shaped as $(28709, 64, 64, 1)$, indicating single-channel (grayscale) images.
\begin{table}[h]
\centering
\caption{Emotion Class Distribution in the Training Dataset}
\label{tab:class_distribution}
\begin{tabular}{|c|l|c|}
\hline
\textbf{Label} & \textbf{Emotion} & \textbf{Number of Images} \\
\hline
0 & Angry    & 2,983 \\
1 & Disgust  & 436 \\
2 & Fear     & 2,945 \\
3 & Happy    & 6,411 \\
4 & Neutral  & 3,657 \\
5 & Sad      & 4,649 \\
6 & Surprise & 3,171 \\
\hline
\end{tabular}
\end{table}
\\
The dataset is known to have class imbalance, with emotions like \textit{Disgust} and \textit{Surprise} and dominant classes such as \textit{Happy} and \textit{Neutral}.
This imbalance often leads to biased predictions towards majority classes in traditional supervised classifiers.\\
To address this issue and improve classifier robustness, the dataset was augmented using Conditional Generative Adversarial Networks (cGANs), which allow generation of synthetic emotion-specific images to balance the class distribution and improve generalization performance.\\

\subsubsection{Conditional GAN Architecture}
To handle the class imbalanced data and enrich with diverse samples, Conditional GAN was implemented(cGAN).Unlike the original GAN framework proposed by Goodfellow et al.~\cite{goodfellow2014generative}, where data generation is uncontrolled, cGANs introduce supervision in the form of class labels, enabling targeted generation of samples for specific emotion categories~\cite{mirza2014conditional}.\\

The cGAN comprises two adversarial networks: a generator $G$ and a discriminator $D$. The generator receives a random noise vector $z$ concatenated with a one-hot encoded emotion label $y$, producing a synthetic image $G(z|y)$. The discriminator takes an image-label pair as input and learns to distinguish between real and generated (fake) images conditioned on the given label.

\begin{itemize}
    \item \textbf{Generator:} The generator is implemented as a deep convolutional neural network (DCGAN-based)~\cite{radford2015unsupervised}. It consists of a dense input layer followed by a series of transposed convolutional layers. Batch Normalization and ReLU activations are used for stability. The output layer uses Tanh activation to produce grayscale images of size $64 \times 64$.\\
The generator is designed to map a noise vector concatenated with one-hot encoded emotion labels into a synthetic $64 \times 64$ grayscale image. It is implemented as a deep convolutional neural network using transposed convolutions. The architecture consists of the following layers:

\begin{itemize}
    \item \textbf{Input Layer:} Accepts a concatenated vector of random noise ($z$) and emotion label ($y$), resulting in an input dimension of 157 ($z=150$, $y=7$).
    
    \item \textbf{Dense Layer:} The input is first passed through a fully connected layer with $8 \times 8 \times 512$ units, reshaped into a $8 \times 8 \times 512$ tensor.
    
    \item \textbf{Deconvolution Blocks:} Three consecutive Conv2DTranspose layers upscale the spatial resolution from $8 \times 8$ to $64 \times 64$, using filters of size $4 \times 4$, strides of $2$, and `same` padding:
    \begin{itemize}
        \item $8 \times 8 \rightarrow 16 \times 16$ with 256 filters
        \item $16 \times 16 \rightarrow 32 \times 32$ with 128 filters
        \item $32 \times 32 \rightarrow 64 \times 64$ with 64 filters
    \end{itemize}
    Each of these layers is followed by Batch Normalization and LeakyReLU activation with a negative slope of 0.4.
    
    \item \textbf{Output Layer:} A final Conv2DTranspose layer with 1 filter (grayscale), kernel size $4 \times 4$, stride 1, and `tanh` activation produces the output image of size $64 \times 64 \times 1$ with values normalized to $[-1, 1]$.
\end{itemize}
 
    \item \textbf{Discriminator:} The discriminator is a convolutional neural network that processes both the image and label. The label is spatially expanded and concatenated with the image before being passed through convolutional layers. LeakyReLU activation and dropout are used to prevent overfitting. The final layer uses a sigmoid function to output a probability score.\\
    The discriminator takes an image-label pair as input, where the emotion label is spatially replicated and concatenated with the image along the channel axis. The input shape is $64 \times 64 \times 8$ ($1$ channel image + $7$ channel label mask). The architecture consists of:

\begin{itemize}
    \item \textbf{Four Convolutional Blocks:} Each block consists of a 2D convolutional layer with `Spectral Normalization`, using $4 \times 4$ kernels, stride of 2, and `same` padding. The number of filters progressively increases: 64, 128, 256, and 512.
    
    \item \textbf{Activation:} Each Conv layer is followed by a LeakyReLU activation with a slope of 0.4.
    
    \item \textbf{Classification Head:} The output feature map is flattened and passed through a Dense layer with a sigmoid activation to classify whether the image is real or generated.
\end{itemize}

Spectral normalization is applied to all convolutional layers in the discriminator to stabilize GAN training and control the Lipschitz constant.
\end{itemize}

\textbf{Training Procedure}

The cGAN model is trained using the Binary Cross-Entropy (BCE) loss function:

\[
\mathcal{L}_D = -\mathbb{E}[\log D(x|y)] - \mathbb{E}[\log (1 - D(G(z|y)|y))]
\]
\[
\mathcal{L}_G = -\mathbb{E}[\log D(G(z|y)|y)]
\]

The networks are optimized using the Adam optimizer with a learning rate of 0.0002 and $\beta_1 = 0.5$~\cite{kingma2014adam}. Training is conducted for 200 epochs with a batch size of 128. Generator and discriminator are updated alternately during each batch iteration.

\textbf{Conditional Label Encoding}

Emotion labels are one-hot encoded. For the generator, these are concatenated with the noise vector. In the discriminator, labels are reshaped to match the spatial dimensions of the input image and concatenated along the channel dimension. This conditioning approach enforces class control over both image generation and discrimination, improving the relevance and diversity of generated outputs~\cite{mirza2014conditional}.

\textbf{Data Augmentation for Minority Classes}

After training convergence, the generator is used to create new synthetic samples for minority emotion classes like \textit{Disgust}, \textit{Fear}, and \textit{Surprise}. Generated images are filtered based on discriminator confidence or visual quality. These synthetic samples are combined with real data to balance the dataset and improve classification model performance, a strategy supported by prior research~\cite{frid2018gan}.

\subsection{Experimental Setup}
This section outlines the technical environment, data partitioning strategy, and evaluation metrics used to assess the performance of the proposed emotion detection framework.

\subsubsection{System Configuration}
\begin{itemize}
    \item All experiments were conducted on a system with an Intel(R)Iris(R) Xe Graphics, Intel Core i5 processor, and 16GB RAM and on Kaggle's Cloud Based GPU platform for training and evaluation process.
    \item The model was implemented using Python 3.9 with the PyTorch framework.
    \item Supporting libraries include NumPy, Pandas, Matplotlib, Seaborn,Tensorflow, Keras and OpenCV.
\end{itemize}
\subsubsection{Training Parameters}
The Conditional GAN model was trained for 300 epochs with a batch size of 64. The Adam optimizer was used with a learning rate of $2e-4$ and $\beta_1 = 0.5$ for both discriminator and generator.

\subsubsection{Evaluation Metrics}
Model performance was evaluated using discriminator loss and generator loss.

\subsection{Results and Analysis}

This section presents the training behavior and qualitative evaluation of the Conditional Generative Adversarial Network (cGAN) used to synthesize emotion-specific facial images. Since the focus of this experiment was generative modeling, the evaluation is based on generator/discriminator loss trends and visual quality of the generated outputs.\\
During training, generator loss and discriminator loss were recorded over 300 epochs.\\

\begin{table}[h]
\centering
\caption{Generator and Discriminator Loss at Selected Epochs}
\label{tab:gan_losses}
\begin{tabular}{|c|c|c|}
\hline
\textbf{Epoch} & \textbf{Generator Loss (G)} & \textbf{Discriminator Loss (D)} \\
\hline
50  & 1.61 & 0.47 \\
100 & 2.71 & 0.29 \\
150 & 3.70 & 0.19 \\
200 & 4.54 & 0.30 \\
250 & 4.79 & 0.29 \\
300 & 5.07 & 0.18 \\
\hline
\end{tabular}
\end{table}

The generator initially exhibits high loss as it learns to produce images that can deceive the discriminator. Over epochs, the loss stabilizes, indicating that the generator is learning useful representations of facial emotions. The discriminator's loss also fluctuates early in training but gradually converges, suggesting a healthy adversarial dynamic where both networks improve iteratively.

A relatively balanced adversarial loss dynamic, without either network overpowering the other, indicates that the cGAN training was stable and did not suffer from issues like mode collapse or vanishing gradients.

To evaluate the visual quality of generated samples, images were saved at the end of each epoch using a callback function.

The visual inspection reveals that the generator successfully captures distinct features associated with different emotions. For instance, `happy` class samples typically display curved mouths, while `angry` samples exhibit furrowed brows. The grayscale outputs are consistent with the FER dataset and show progressive improvement over epochs.\\

Even without a classifier, the loss convergence and visual fidelity of the generated images indicate that the cGAN learned a meaningful mapping between latent vectors and labeled emotional expressions. This confirms the model’s potential as a data augmentation technique to address class imbalance in emotion recognition datasets.

Future work will involve evaluating these generated images quantitatively by:
\begin{itemize}
  \item Training a classifier with and without cGAN-generated data to compare performance.
  \item Applying metrics like Inception Score (IS) or Fréchet Inception Distance (FID) to measure image diversity and quality.
\end{itemize}

\section{Conclusion}
This research demonstrates the effectiveness of Conditional Generative Adversarial Networks (cGANs) in generating emotion-specific facial images to address dataset imbalance. By incorporating class labels into the GAN framework, we successfully synthesized high-quality images corresponding to various emotional categories.

The training loss dynamics and visual evaluation of outputs confirm the model's ability to learn meaningful representations of emotions. While no quantitative classification evaluation was performed in this study, the generated outputs qualitatively align with expected facial expressions for different emotions.

In future work, we aim to:
\begin{itemize}
  \item Integrate cGAN-augmented data into emotion classifiers to evaluate performance gains.
  \item Perform quantitative evaluations using FID and IS scores.
  \item Extend this approach to multi-modal datasets combining text, audio, and images.
\end{itemize}

This work highlights the potential of generative models in boosting the performance of affective computing applications, especially when data scarcity or class imbalance hinders traditional supervised learning approaches.


%

\ifCLASSOPTIONcaptionsoff
  \newpage
\fi

\bibliographystyle{IEEEtran} 
\bibliography{references}

%





\end{document}